# A Food Package Recognition and Sorting System Based on Structured Light and Deep Learning

Xuanzhi Liu

Shenzhen Institute of Advanced Technology, Chinese Academy of Sciences; Guangdong University of Finance & Economics, xuanzhibill@gmail.com

Jixin Liang

Shenzhen Institute of Advanced Technology, Chinese Academy of Sciences, jx.liang2@siat.ac.cn

Yuping Ye

Shenzhen Institute of Advanced Technology, Chinese Academy of Sciences, yp.ye@siat.ac.cn

Zhan Song*

Shenzhen Institute of Advanced Technology, Chinese Academy of Sciences, zhan.song@siat.ac.cn

Juan Zhao

Shenzhen Institute of Advanced Technology, Chinese Academy of Sciences, juan.zhao@siat.ac.cn

Vision algorithm-based robotic arm grasping system is one of the robotic arm systems that can be applied to a wide range of scenarios. It uses algorithms to automatically identify the location of the target and guide the robotic arm to grasp it, which has more flexible features than the teachable robotic arm grasping system. However, for some food packages, their transparent packages or reflective materials bring challenges to the recognition of vision algorithms, and traditional vision algorithms cannot achieve high accuracy for these packages. In addition, in the process of robotic arm grasping, the positioning on the z-axis height still requires manual setting of parameters, which may cause errors. Based on the above two problems, we designed a sorting system for food packaging using deep learning algorithms and structured light 3D reconstruction technology. Using a pre-trained MASK R-CNN model to recognize the class of the object in the image and get its 2D coordinates, then using structured light 3D reconstruction technique to calculate its 3D coordinates, and finally after the coordinate system conversion to guide the robotic arm for grasping. After testing, it is shown that the method can fully automate the recognition and grasping of different kinds of food packages with high accuracy. Using this method, it can help food manufacturers to reduce production costs and improve production efficiency.

CCS CONCEPTS • Computing methodologies~Artificial intelligence~Computer vision~Computer vision tasks~Vision for robotics

---

* Corresponding author: zhan.song@siat.ac.cn

## 1 INTRODUCTION

With the support of modern science and technology, the use of industrial robots is gradually gaining popularity. They have the advantages of reliability, stability, and high precision, which can reduce the intensity of manual work and thus improve the quality of work based on the guarantee of operational efficiency. As a result of these advantages, industrial robots have gradually developed into a core force in the manufacturing industry [1-3]. Existing industrial robots usually work according to a preset position, and once the position changes, they cannot continue to work and must be reprogrammed [4]. Along with the deepening development of artificial intelligence, ChatGPT and other intelligent machine technologies, intelligent machines are continuously reshaping the manufacturing system [5]. In recent years, computer vision algorithms have been widely used in robotics, object recognition, and other fields, and some recent studies have applied computer vision to robotic arm grasping tasks [6-8]. Yao et al. [9] designed a robotic arm intelligent grasping system based on machine vision, using a matching algorithm based on contour features to identify target objects and capable of completing grasping and handling tasks. Liu et al. [10] designed a vision-based mobile sorting robot system, which also uses image binarization, edge detection and other algorithms to determine the location of the target and then guide the robot to grasp it.

However, for the sorting of food packages, the existing technology suffers from the following two problems. The first is the lack of robustness of traditional vision algorithms [11]. Food packaging usually uses some reflective or transparent materials, which can reflect or transmit light, making many pixel points in the image captured by the camera overexposed or too dark, and the parameters in the traditional vision algorithm cannot fit these pixel points well. In addition, there is a lack of a simple, automatic method to determine the height of the target on the z-axis in guiding the robot arm for grasping. In order to solve the above two problems, we designed a fully automatic food package recognition and sorting system to achieve a high accuracy rate of sorting for food packages.

Figure 1 shows the whole system and the objects we want to detect. In Figure 1(a), the depth camera is located at a height of 860 mm directly above the target, and the initial position of the robotic arm is located outside the field of view of the camera. Figure 1(b) shows the three kinds of food packages we used. To increase the difficulty of identification, we chose three packages with different degrees of reflectivity and transparency, and placed the packages in a random stack.

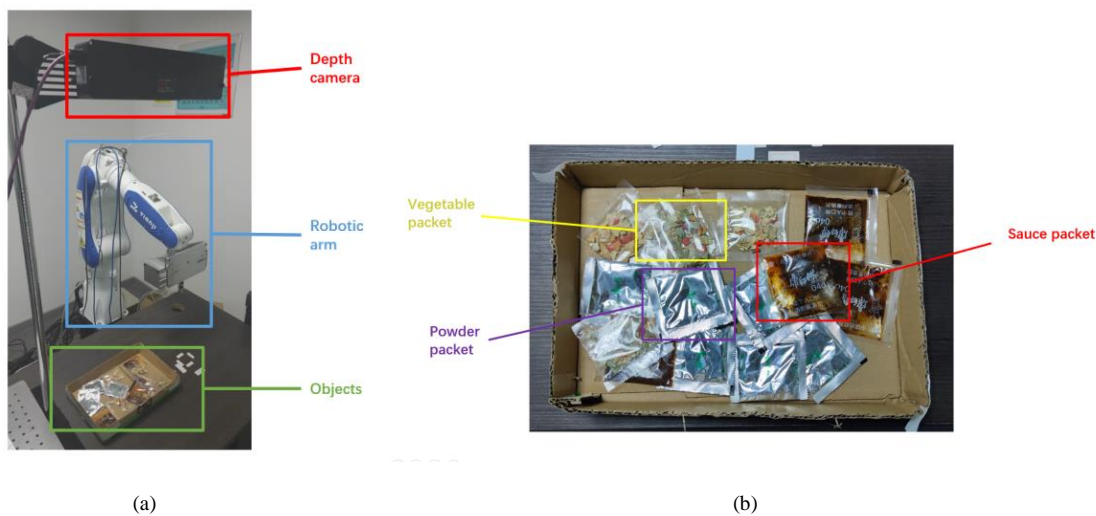

(a) (b)

Figure 1: The system we designed (a) and three kinds of packages we detected (b).

The difficulty of identification lies in the detection of powder packets and vegetable packets. These two types of packaging are highly reflective characteristics, may prone to errors and omissions after random stack.

## 2 METHODOLOGY

In Figure 1, we show the system we designed and three kinds of food packages we used for testing. The detailed workflow of the system is shown in Figure 2, which consists of three modules: structured light module, deep learning module and grasping module. Initially, the target within the field of view is scanned using a depth camera to generate a 3D point cloud, while the 2D image is transmitted to the deep learning module. In this module, the pre-trained MASK R-CNN model is utilized to classify the target's class and provide the predicted 2D coordinates to the point cloud reconstruction module. Subsequently, based on the reconstruction outcomes, the 3D coordinates corresponding to the 2D points in the point cloud are determined. Ultimately, the coordinates are transformed through hand-eye calibration and computed as points within the robot arm coordinate system for the purpose of grasping.

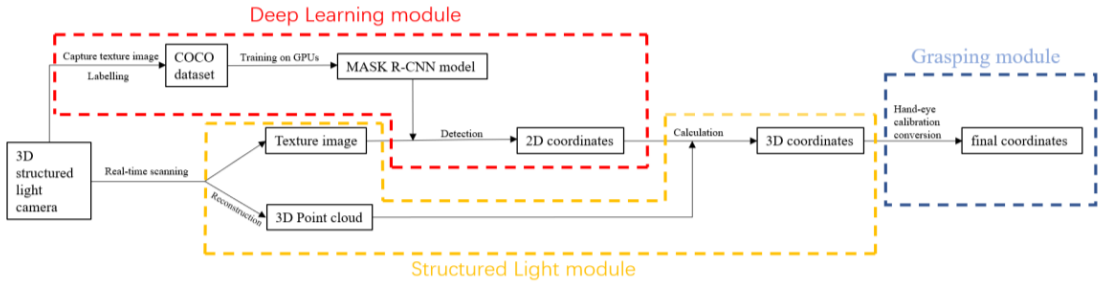

Figure 2: Framework of the system.

### 2.1 Structured light module

In order to reconstruct the 3D information of the target while obtaining a 2D image of the target, we use a structured light-based depth camera to scan the target. By using the projection equipment to project a pre-designed structured light pattern of a specific pattern onto the surface of the object, and then using the camera to photograph the deformation pattern, the 3D information of the object to be measured is obtained by analyzing the aberrations produced by the projected pattern on the surface of the object under study, and then using the specific encoding scheme of the projection to obtain the three-dimensional coordinates of the points on the target to be measured using the triangulation principle of the pinhole imaging pattern of the camera. In this system, the encoded stripe pattern we use is an 18-sheet stripe pattern made by Gray code and line shift code [12]. Figure 3 shows the mathematical model of the 3D reconstruction. $M$ is a point on the object to be measured, then $M$ in the projector image corresponds to the point $m_p = (u_p, v_p)^T$, in the camera image corresponds to the point $m_c = (u_c, v_c)^T$. Figure 4 shows the results of the 3D reconstruction in the form of a depth map, the colored pixel points represent the successfully reconstructed 3D points, and the black points represent the missing 3D points.

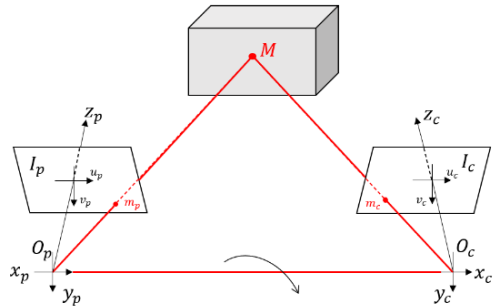

Figure 3: Mathematical model for structured light 3D reconstruction.

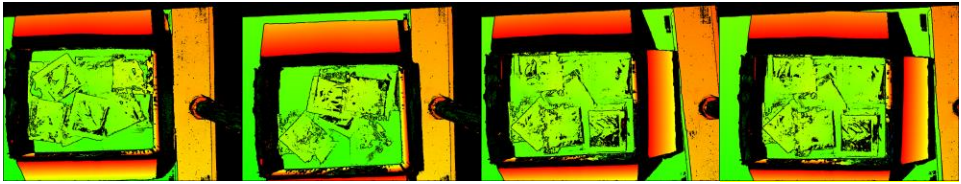

Figure 4: The results of the partial 3D reconstruction are presented as a depth map.

Once the 2D coordinate information of the target $P_{2D}(X_{2D}, y_{2D})$ is obtained, its corresponding 3D coordinate $P_{3D}(X_{3D}, y_{3D}, z_{3D})$ can be calculated as follows:

$$index = y_{2D} W_{pic} + x_{2D} \quad (1)$$

$$x_{3D} = x[index] \quad (2)$$

$$y_{3D} = y[index] \quad (3)$$

$$z_{3D} = z[index] \quad (4)$$

where $index$ is the index of the 2D coordinate, $W_{pic}$ is the width of the 2D image, and $x$, $y$, and $z$ store the 3D point cloud data of the image.

## 2.2 Deep learning module

With the introduction of convolutional neural networks and the continuous development of deep learning methods in the field of computer vision, object detection methods based on convolutional neural networks have gradually emerged and replaced traditional methods as the mainstream. Referring to the research results of Luo's team [13], Yuan's team [14] and Wang's team [15], we used the MASK R-CNN [16] algorithm with deep learning instance segmentation instead of the traditional vision algorithm for food packaging detection of reflective and transparent materials. Figure 5 shows the flow of the algorithm. It uses ResNet [17] as CNN [18] backbone and extends the Faster R-CNN [19] by incorporating a segmentation branch for generating object masks. The input image is first input into a CNN for image feature extraction to obtain a feature map, which is usually ResNet50 or ResNet101. Then the RPN (Region Proposal Network) is used to generate regions that may contain targets, and for each region, it is projected onto the corresponding feature map using ROI Align to generate a fixed size feature map. Finally, the feature map is fed into a Full Convolutional Network (FCN), which is used to segment the instances at the pixel level, generate a mask for each instance, and finally output the class, location and mask information for each instance. By deploying the model in the system and reading the grayscale image obtained from the depth camera scan, the 2D position information and class information of the target in the image can be predicted.

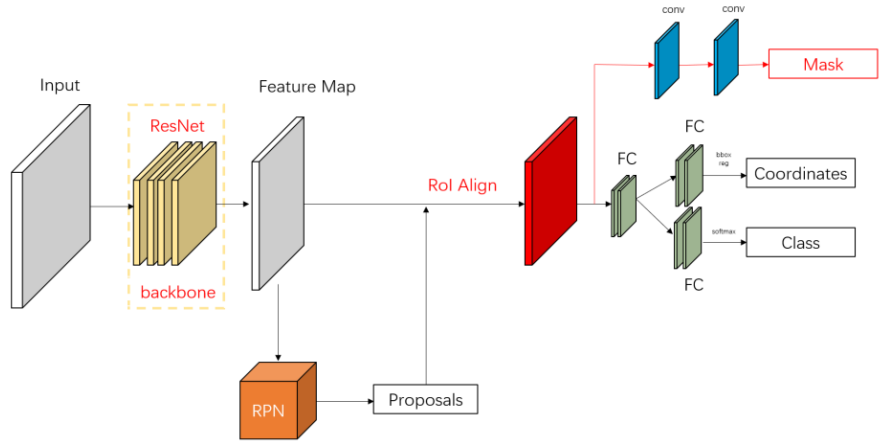

Figure 5: Flowchart of MASK R-CNN

For the calculation of 2D coordinates, we calculate its midpoint as the center of the object based on the predicted coordinates of the top-left and bottom-right corners of the rectangular box. Since the 3D coordinates of this center point may be missing, for each target, we take a small area near its midpoint, calculate all the 3D coordinates corresponding to the points in this area, and then sort them according to the z magnitude, and select the one with the smallest z as the final result to input into the next module.

Since MASK R-CNN is a supervised learning network, its algorithm needs to be trained by using datasets with labels for the purpose of accurately classifying data or predicting results. In this system, we also use a depth camera to capture grayscale images of objects, label them as a dataset and then feed the dataset into a MASK R-CNN network model to be trained in a GPU environment. We choose grayscale images as the input to the network model because they have better morphological characteristics than RGB images [20] and have only one channel, which is easy to transmit and compute in the system. Figure 6 shows part of the dataset after the labeling process. The colored parts of the image are manually labeled with different instance labels, and this label information will be used for network model training along with the pixel information of the original image.

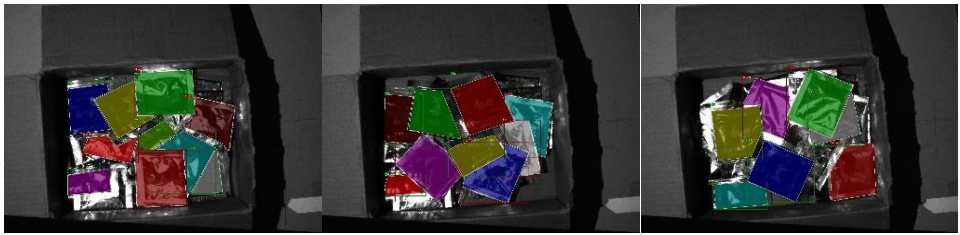

Figure 6: Some of the images after the tagging process.

## 2.3 Grasping module

In structured light module, the 3D coordinates obtained are based on the camera coordinate system and need to be converted to the robot arm base coordinate system in order to control the robot arm to that point for gripping.

Before the coordinate conversion, the hand-eye calibration of the robot arm is needed to obtain the relative positional relationship between the robot arm and the camera. We use a calibration plate for hand-eye calibration. An algorithm is used to identify the corner points in the calibration plate, then the end-effector of the robot arm is manually controlled to the position of the corner points and the conversion between them is calculated based on the coordinates of the end-effector and the coordinates of the corner points identified by the camera.

After the calibration is completed, the coordinates of the points under the camera coordinate system can be calculated corresponding to the coordinates under the robot arm base coordinate system. For the calculated 3D point coordinates $P_{3D}(x_{3D}, y_{3D}, z_{3D})$ under the coordinate system $P_{camera}$, its corresponding point under the robot base coordinate system $P_{robot}$ can be calculated by a transformation matrix $H$:

$$H = \begin{bmatrix} R & T \\ O & 1 \end{bmatrix} \quad (5)$$

where $R$ is called the rotation matrix and $T$ is called the translation vector.

Once the transformation matrix $H$ has been constructed, the corresponding values for the points under $P_{camera}$ can be calculated under $P_{camera}$

$$\begin{bmatrix} P_{Robot} \\ 1 \end{bmatrix} = H * \begin{bmatrix} P_{Camera} \\ 1 \end{bmatrix} \quad (6)$$

That is, for a point $P_{ij}$ in the $P_{camera}$ coordinate system, there is:

$$P_{ij} = \begin{bmatrix} x_{ij} \\ y_{ij} \\ z_{ij} \end{bmatrix} \quad (7)$$

$$P'_{ij} = R * P_{ij} + T \quad (8)$$

where $P'_{ij}$ is the corresponding point of $P_{ij}$ point under the $P_{robot}$ coordinate system.

Input $P'_{ij}$ as the final result into the robot arm, and then the robot arm can be commanded to run to that point for grasping.

## 3 EXPERIMENTAL RESULTS

### 3.1 Model training results

We divided the 300 dataset images into a training set and a validation set in the ratio of 7:3. Each image is 2044*1536 in a single-channel grayscale image. A server with eight GeForce RTX 3090 graphics cards was used for the training experiments. There are a number of hyperparameters that need to be set in the training of the model. We set the batch-size of the network to 16, meaning that two images can be read at once on each GPU. In back propagation, the SGD optimizer was chosen for training, with an initial learning rate set to 0.02, a momentum of 0.9, a weight decay of 0.0001, and a maximum epoch of 100. A strategy of linear growth of the initial learning rate was used. At the beginning of training, the learning rate starts at 0.001 and gradually increases to 0.02 through 500 iterations and is multiplied by 10% at the 80th and 90th epochs. ResNet101 was chosen for the backbone part of the network, num classes were set to 3 (without background).

The change in loss and the improvement of the accuracy graph during the training process are shown in Figure 7. The loss finally converges at about 0.06 and the accuracy curve eventually stabilizes at about 99.63.

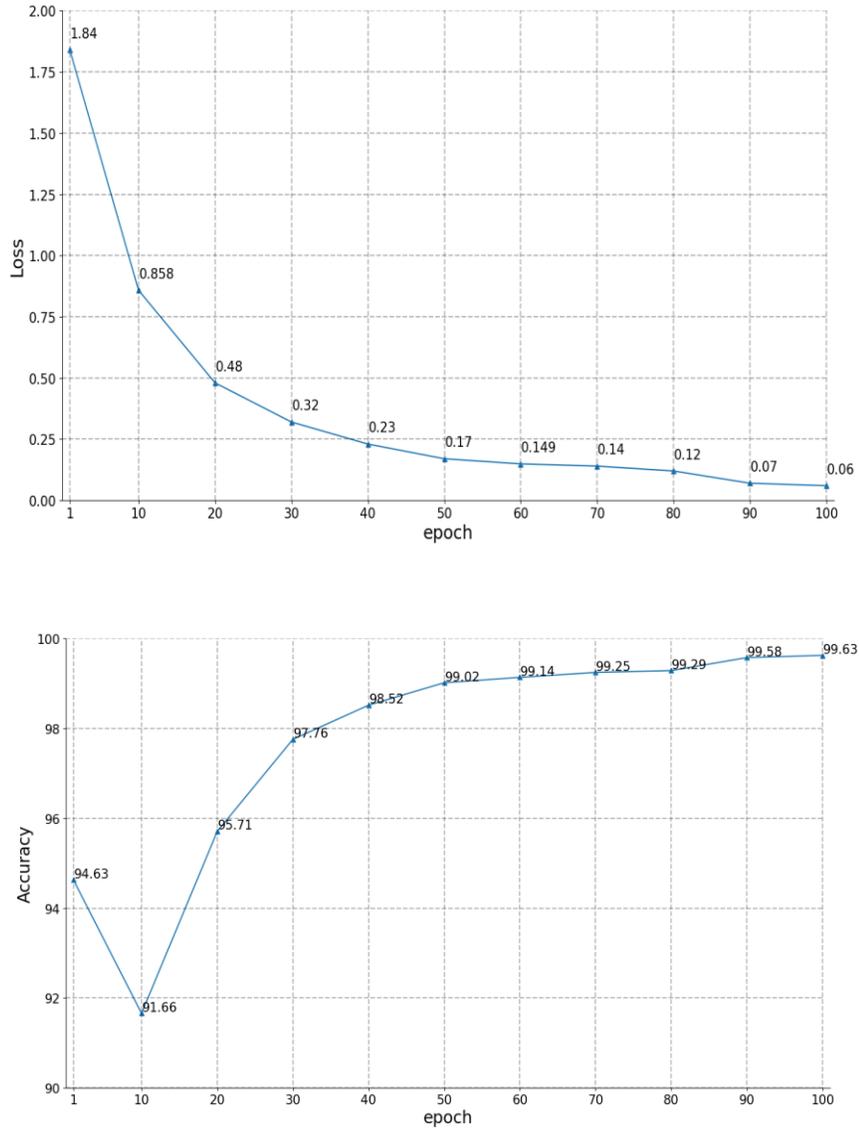

Figure 7: The loss curve and accuracy curve.

## 3.2 Testing the model on a test set

We tested the model using images from the validation set. We set the threshold for model prediction to 0.8 because we found that for the targets we expect to be predicted, the probability of the prediction result is between 0.8 and 1, while for some targets that we do not expect to be predicted, the probability of their prediction result is mostly below 0.8. The mask of such targets usually has a large error or the location is not favorable for grasping, so we hope to reduce the uncertainty in the crawl by setting a higher threshold. Figure 8 shows the prediction results of the model.

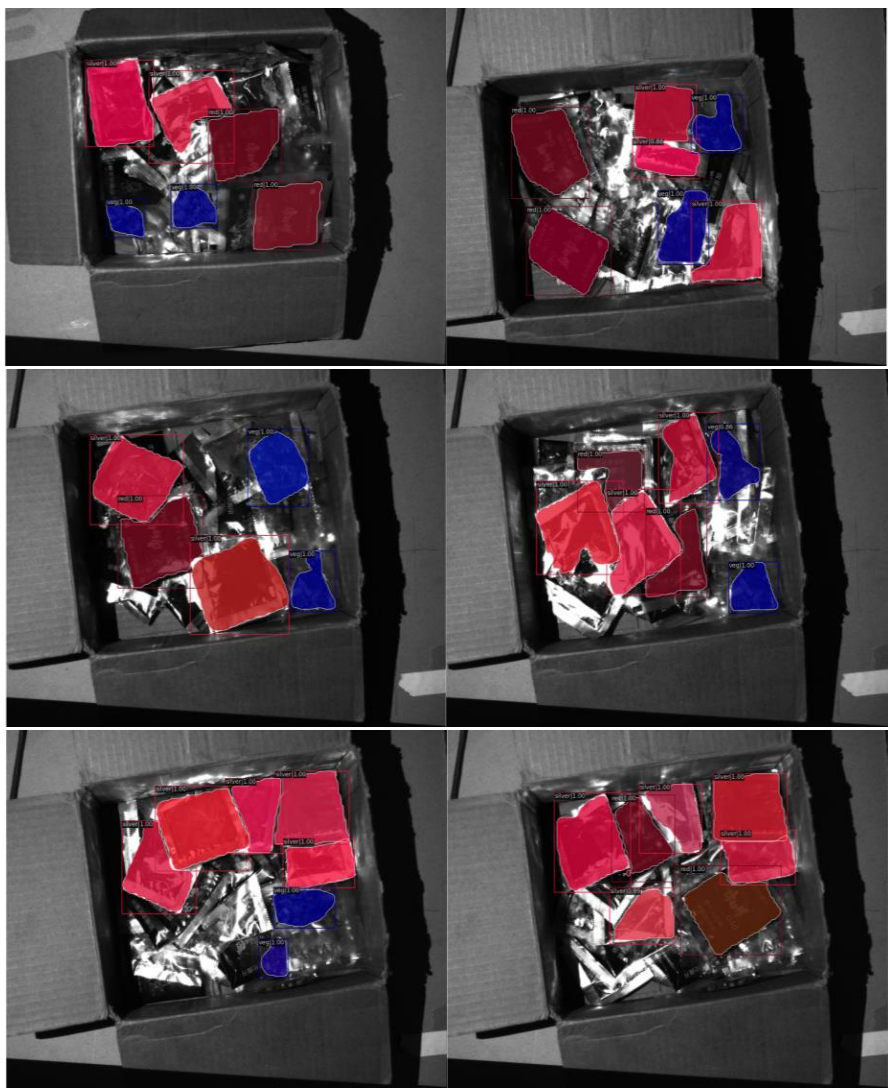

Figure 8: Some prediction results of MASK R-CNN model. A 100% correct rate was achieved, with no false checks.

### 3.3 Final test results

Tests have shown that the method can achieve automatic recognition and grasping for a wide range of food packages. Figure 9 shows one set of test results, where we placed four sauce packages, four powder packages and five vegetable packages. It first goes through one scan, then grabs a vegetable packet, then a second scan, grabs a sauce packet, then a third scan, grabs a powder packet. The process is cycled through until there are no targets in the field of view. The system finally achieved 100% correct grasping rate.

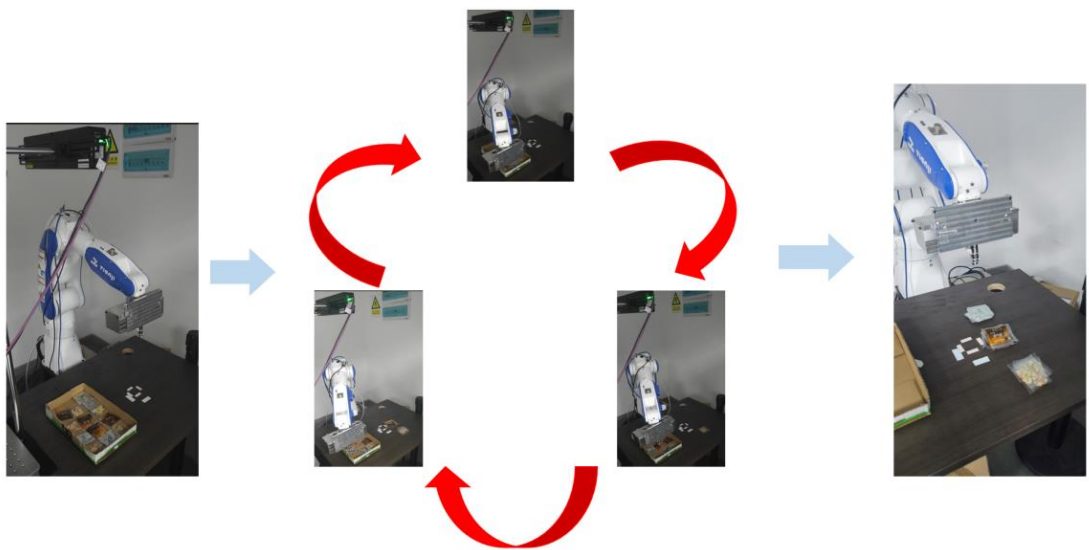

Figure 9: The system's operation process.

## 4 CONCLUSION AND FUTURE WORK

Through experiments, we demonstrated that training and prediction using the MASK R-CNN network model can solve the recognition problem of reflective and transparent packaging. This method can identify most of the obvious targets on the surface, and then calculate the two-dimensional results into three-dimensional ones based on the point cloud, which can improve the grasping accuracy in the z-axis direction. However, as our training dataset is small, there is still much room for improvement in model prediction, in the future we are considering expanding a larger dataset for model training. In addition, we realize the importance of polarization optical imaging techniques in improving object imaging quality and enhancing identification probability based on the difference in polarization characteristics of reflected light from the object and stray light from the background. It can remove strong specular reflections from highly reflective material surfaces, thus improving the imaging quality of camera scans. In the subsequent work, we will try to modify the depth camera using polarized lenses, which can improve the quality of the point cloud after 3D reconstruction while improving the probability of object detection.

## 5 ACKNOWLEDGMENT

This work was supported in part by Key-Area Research and Development Program of Guangdong Province, China (2019B010149002) and Shenzhen Science and Technology Program (JSGG20220831092801003).
## REFERENCES

[1] Dai Haozen, Sun Daning, Zhao Wenbo. A review of the development and application of industrial robotics[J]. The Journal of New Industrialization,2021,11(04):5-6.DOI:10.19335/j.cnki.2095-6649.2021.4.003.
[2] Feng X, Song M, Ni XY, et al. Review of industrial robot development[J]. Technology innovation and Application,2019,No.280(24):52-54.
[3] Meng M. F., Zhou T. D., Chen L. B. et al. A review of the Research and Development of Industrial robots[J]. Journal of Shanghai Jiaotong University,2016,50(S1):98-101.DOI:10.16183/j.cnki.jsjtu.2016.S.025.
[4] Jin Hanlin. Research on Target Grasping for Manipulators Based on Visions [J]. Mechanical Engineering and Automation,2018,No.209(04):33-34+37.
[5] He Jiang, Yan Shumin, Zhu Li Li et al. Analysis of Interaction Relationship between Industrial Robots and Human Workforce[J]. Shanghai Journal of Economics,2023,No.414(03):71-87.
[6] Cai Chen,Wei Guoliang. An improved Manipulator Grasping Method Based on Convolutional Neural Network[J]. Computer and Digital Engineering,2020,48(01):158-162.
[7] Xia Qunfeng,Peng Yonggang. Review on application research of robots scraping system based on visual[J]. Journal of Mechanical & Electrical Engineering,2014,31(06):697-701+710.
[8] WANG Chengjun,WEI Zhiwen,YAN Chen. Review on Sorting Robot Research Based on Machine Vision Technology[J]. Science Technology and Engineering,2022,22(03):893-902.
[9] Yao Qicai, Wang Di, Liao Maosheng. Intelligent Grasping System Design of Robot Arm Based on Machine Vision[J].Metrology and Measurement Technique,2020,47(10):28-33.DOI:10.15988/j.cnki.1004-6941.2020.10.009.
[10] Liu Xiao, Wan Qi, Ni Yintang et al. Research on Sorting Robot System Based on Vision for Mobile Sorting[J]. Industrial Control Computer,2018,31(12):91-93.
[11] Li Tao-tao. Research on License Plate Recognition Based on Convolutional Neural Network[J]. Agricultural Equipment and Vehicle Engineering,2021,59(05):119-121.
[12] Song Z, Chung R, Zhang X T. An accurate and robust strip-edge-based structured light means for shiny surface micromeasurement in 3-D[J]. IEEE Transactions on Industrial Electronics, 2012, 60(3): 1023-1032.
[13] Luo Jing,Chen Jinhai,Peng Zhixuan et al. Robot grasping experimental system based on machine vision[J]. Experimental Technology and Management,2022,39(04):45-50.DOI:10.16791/j.cnki.sjg.2022.04.010.
[14] Yuan Yuan,Chen Yu,Zhou Qinghua et al. Visual Grasping method of rope-drive manipulator using IMask R-CNN[J]. Computer Application Research,2021,38(10):3093-3097.DOI:10.19734/j.issn.1001-3695.2021.03.0082.
[15] WANG Deming,YAN Yi,ZHOU Guangliang et al. 3D Vision-Based Picking System with Instance Segmentation Network and Iterative Optimization Method[J]. ROBOT,2019,41(05):637-648.DOI:10.13973/j.cnki.robot.180806.
[16] He K, Gkioxari G, Dollár P, et al. Mask r-cnn[C]//Proceedings of the IEEE international conference on computer vision. 2017: 2961-2969.
[17] He K, Zhang X, Ren S, et al. Deep residual learning for image recognition[C]//Proceedings of the IEEE conference on computer vision and pattern recognition. 2016: 770-778.
[18] Krizhevsky A, Sutskever I, Hinton G E. Imagenet classification with deep convolutional neural networks[J]. Communications of the ACM, 2017, 60(6): 84-90.
[19] Ren S, He K, Girshick R, et al. Faster r-cnn: Towards real-time object detection with region proposal networks[J]. Advances in neural information processing systems, 2015, 28.
[20] Yang Gui.Study on River Flow Measurement Based on CNN and Image Processing[D]. Shan dong University, 2021. DOI:10.27272/d.cnki.gshdu.2021.004501.
10